\documentclass{article}
\usepackage{spconf,amsmath,graphicx,amsfonts,xcolor,url,booktabs,multirow,subcaption,comment,siunitx}
\usepackage[inline]{enumitem}

\usepackage{algorithm}
\usepackage[noend]{algpseudocode}

\newcommand*{\defeq}{\stackrel{\text{def}}{=}}

\title{Federated Transfer Learning with Dynamic Gradient Aggregation}

\name{Dimitrios Dimitriadis, Kenichi Kumatani, Robert Gmyr, Yashesh Gaur and Sefik Emre Eskimez}
\address{Microsoft, Redmond, WA, USA}

\begin{document}
\ninept
\maketitle
\begin{abstract}
In this paper, a Federated Learning (FL) simulation platform is introduced. The target scenario is Acoustic Model training based on this platform. To our knowledge, this is the first attempt to apply FL techniques to Speech Recognition tasks due to the inherent complexity. 
The proposed FL platform can support different tasks based on the adopted modular design. As part of the platform, a novel hierarchical optimization scheme and two gradient aggregation methods are proposed, leading to almost an order of magnitude improvement in training convergence speed compared to other distributed or FL training algorithms like BMUF and FedAvg. 
The hierarchical optimization offers additional flexibility in the training pipeline besides the enhanced convergence speed. On top of the hierarchical optimization, a dynamic gradient aggregation algorithm is proposed, based on a data-driven weight inference. This aggregation algorithm acts as a regularizer of the gradient quality. 
Finally, an unsupervised training pipeline tailored to FL is presented as a separate training scenario.
The experimental validation of the proposed system is based on two tasks: first,  the LibriSpeech task showing a speed-up of $7\times$ and $6\%$ word error rate reduction (WERR) compared to the baseline results. The second task is based on session adaptation providing an improvement of $20\%$ WERR over a competitive production-ready LAS model. The proposed Federated Learning system is shown to outperform the golden standard of distributed training in both convergence speed and overall model performance.

\end{abstract}

\begin{keywords}
Acoustic Modeling, distributed training, federated learning
\end{keywords}

\section{Introduction}
\label{sec:intro}

Distributed Training (DT) is drawing much attention with the goal of scaling the model training processes. Since the training datasets become ever larger, the need for training parallelization becomes more pressing. Different approaches have been proposed over the years~\cite{BNH19}, aiming at more efficient training, either in the form of training platforms such as ``Horovod''~\cite{SeBa18, Abadi+16} or algorithmic improvements like ``Blockwise Model-Update Filtering'' (BMUF)~\cite{chen2016scalable}. These techniques are evaluated on metrics such as data throughput (without compromising accuracy), model and/or training dataset size, and GPU utilization. However, a few underlying assumptions are implied as part of such DT scenarios, i.e., data and device uniformity and efficient network communication between the working nodes. Besides the communication/network specifications (not examined in this paper), the data uniformity is paramount for the successful training, ensured by repeated randomization and data shuffling steps.

Meanwhile, new constraints in data management are emerging nowadays. Some of these constraints are driven by the need for privacy compliance of the personal data and information~\cite{GDPR}. As such, increasingly more data is stored behind inaccessible firewalls or on users' devices without the option of sharing for centralized training. To this end, the Federated Learning (FL) paradigm has been proposed, addressing the privacy concerns, while still processing such inaccessible data. The proposed approach aims at training ML models, e.g., deep neural networks, on data found on multiples of local worker nodes without the need to exchange any data between the ``coordinator'' and these remote nodes. The general principle is based on training different versions of the model on that local data samples, while exchanging only updates of the model parameters, such as the network parameters or the corresponding gradients. An additional step of synchronizing these local models and updating the global model at an appropriate frequency is now required. More details about general FL techniques can be found in~\cite{Li+20}. Federated Learning is mostly focused on communication efficiency, better optimization~\cite{SSZ13} and/or privacy aspects. 
There are different approaches for FL using either a central server,~\cite{PaY09}, i.e., a ``coordinator or orchestrator,'' or employing peer-to-peer learning, without using a central server,~\cite{LJSC20} -- herein, the first approach is followed. A single server is responsible for the sampling and communication between the clients, updating the models, and adjusting the learning rate. 

Other differences between FL and DT lie on the assumptions made about the properties of the local data sets~\cite{KMR15}. DT primarily aims at parallelizing local computing power, whereas FL focuses on training with heterogeneous data sets. Since the DT focus is the training of a single model on multiple nodes, a common underlying constraint is that all local data subsets need to be as homogeneous as possible, i.e., uniformly distributed and roughly about the same size. However, none of these constraints are necessary for FL; instead, the data sets are typically heterogeneous, and their size may span several orders of magnitude. The FL provides a more flexible training framework, relaxing most of the DT constraints.

Algorithms for FL are designed for model training featuring data parallelism across a large number $K$ of nodes, data imbalance, and data sparseness of local training examples,~\cite{KMR15}.
However, it is possible that the data found in each of the clients can be skewed towards different distributions -- especially in SR applications, where accented speech, background noise, or other factors can have an adverse outcome. Therefore, the FL algorithms need to consider such a challenge. Although we are not, herein, investigating the data sparsity challenge explicitly, the proposed algorithms on the optimization side of FL can implicitly address the data diversity issue.

To the best of our knowledge, a massively distributed and heterogeneous approach, like the one herein presented for Automatic Speech Recognition (ASR), has not been applied before -- albeit, some work exists for KWS~\cite{Leroy+19}. An end-to-end (e2e) architecture is implemented on the Federated Learning platform for this particular SR task. Training of such all-neural models is much simpler than training conventional SR systems, and as such, it is easier to automate. The seq2seq models have gained in popularity in SR tasks because acoustic, language and pronunciation models of a conventional ASR system can be combined into a single neural network~\cite{Chiu+18}. There has been a variety of models proposed, including ``Recurrent Neural Network Transducer'' (RNN-T)~\cite{Graves12}, ``Listen, Attend and Spell'' (LAS)~\cite{CJLV15} and others.  Herein, the LAS architecture is adopted because it consistently provides the best offline results in our internal test sets. The ``seq2seq with attention'' model includes an encoder (similar to the traditional acoustic model), an attention layer, and a decoder (like the language model). More details can be found in Section~\ref{subsec:las}.

The contributions of the paper can be outlined as follows:
\begin{enumerate*}[label=\roman*.]
\item Optimization algorithm: The proposed hierarchical optimization scheme significantly speeds up convergence speed and improves overall classification performance. The ``Generalized FedAvg,''~\cite{Reddi+20} and ``BMUF,''~\cite{chen2016scalable} algorithms overlap with the proposed method.
\item Dynamic Gradient Aggregation: A novel algorithm for self-cleansing batches of ``bad'' data is presented. Similar algorithms have been investigating in~\cite{Alain+16,BTPG16} for the case of ``Asynchronous SGD.'' However, the proposed approach is applied on an FL setting with a data-driven method.
\item Unsupervised Adaptation: An algorithm for unsupervised adaptation of ASR models as part of the FL platform is presented. Unsupervised training using TTS has been proposed before in~\cite{Rosenberg+19}, but it has not been applied on a session adaptation scenario like this one in FTL, and 
\item Novelty of the task: The FL approach for ASR model training is investigated for the first time.
\end{enumerate*}

The paper is focused on the Federated Learning platform, called here \emph{``Federated Transfer Learning''} (FTL platform), for the SR task. Apart from the platform description, algorithmic innovations herein presented are the use of hierarchical optimization, the weighted model aggregation, and the use of unsupervised strategies for SR tasks. The paper consists of the following sections:
\begin{enumerate*}[label=\roman*.]
\item In Section~\ref{sec:background}, an overview of the current state-of-the-art in FL is provided. The overview description is not task-specific. 
\item In Section~\ref{sec:proposal}, more theoretical justification, and details about the SR system are provided. Since this is the first application of FL in SR, theoretical extensions to this particular task are discussed. 
\item Two optimization algorithms are provided for hierarchical optimization and adaptive gradient aggregation in Sections~\ref{sec:proposal} and~\ref{sec:aggregate}.
\item The experimental results are presented in Section~\ref{sec:experiments}, and 
\item Finally, discussion and conclusions can be found in Section~\ref{sec:conclusion}.
\end{enumerate*}
The proposed system shows improvements in convergence speed and classification performance (in terms of Word Error Rates (WER)) for the supervised task and 20\% relative improvements in WER for the unsupervised task.

\section{Background}
\label{sec:background}

\subsection{Federated Learning Background}

Training statistical models using traditional distributed learning algorithms on real application data $\{x_i, y_i\}, i=1,\dots, N$ requires the following steps: copying the data on a centralized storage location, shuffling, evenly distributing and then, training the models with them, similarly to~\cite{KMR15}.  On the other hand, the proposed FL approach follows a different data-handling paradigm, requiring minimal data transfers, thus enhancing privacy. The training constraints are more relaxed since the available computing nodes can be diverse or even inaccessible for periods of time. The proposed algorithms do not require sweeping of the entire data set (across the network of clients) for training but rather just sampling the available nodes in every given iteration. As an additional task, unsupervised training is investigated with the constraint of labeled data alleviated. Herein, an unsupervised SR model training method is proposed as part of the FTL platform.

\subsection{Unsupervised Model Adaptation}
\label{subsec:unsupervised}
The state-of-the-art approach for learning without labels in the traditional centralized setup is based on the semi-supervised training method, as in~\cite{HWG16,LiSWZG17,MosnerWRPKSMH19}, while using a sophisticated model as the teacher to produce targets $y_i$ for the untranscribed data $x_i$. However, this approach is not efficient for FL applications because of the lack of realistic quality control, increased computational complexity, and other FL-related limitations. It is not feasible to transmit and update multiple models, such as the student and ``heavy'' teacher models, especially in poor network conditions and low computational power devices. Herein, we propose using a self-supervised adaptation algorithm,~\cite{Sa94}, based on audio created with a TTS engine,~\cite{Rosenberg+19}, and text found on each client. As such, only the seed model is adapted with data locally produced. The adapted models are then aggregated as part of the FL process. However, the TTS-based audio deteriorates the performance of the encoder after adaptation due to overfitting. To alleviate the adverse effects of synthetic speech, we propose using real speech on the server-side training step, as shown in Figure~\ref{fig:FTL_platform}, to regularize the adaptation process. The combination of TTS-based audio and real speech is unique on the FTL training setup while significantly improving the quality of the adapted models. The regularization step held on the server can also be seen as another method against Catastrophic Forgetting,~\cite{HLRK18, Parisi+18}, sharing similarities with ``Naive Rehearsal'' techniques.

\subsection{Attention-based Sequence-to-Sequence models (seq2seq)}
\label{subsec:las}

Attention-based sequence-to-sequence (seq2seq) models are shown to yield state-of-the-art performance for various ASR tasks~\cite{CJLV15}. A seq2seq model is composed of 3 sub-networks: encoder, decoder and attention. Given speech input $X=\{x_1,...,x_T\}$, the encoder first converts it into a sequence of high-level representations $H^{enc}$,
\begin{align}
    H^{enc} &=\{h^{enc}_1,...,h^{enc}_T\}={\rm Encoder}(X).  
 \label{eq:enc} 
\end{align}
Herein, the encoder is implemented by a bidirectional LSTM~\cite{GFS15} with LayerNorm~\cite{BKH16}.

The decoder acts as an acoustically-conditioned Language Model. For predicting a certain output token $y_n$, the acoustic signal to be used for conditioning is summarized by the attention module. For every  time-step $n$ of the decoder, attention generates alignments $\alpha_n$ over $H^{enc}$, and a corresponding context vector $c_n$. The attention layer is implemented as location-aware attention,~\cite{Chorowski+15}. 
\begin{align}
 c_n, \alpha_n &= {\rm LocationAwareAttention}(d_n, \alpha_{n-1}, H^{enc}). \label{eq:att} 
\end{align}
Here, $d_n$ is the decoder state vector at time $n$. The context vector, $c_n$, is leveraged by the decoder as,
\begin{align}
    d_n, h^{dec}_n &={\rm Decoder}(y_{n-1}, c_{n-1}, h^{dec}_{n-1}), \\
    y_n &= {\rm DecoderOut}(c_n,d_n). 
 \label{eq:dec}
\end{align}
where ${\mathbf Y}=\{y_n\}$ is the output hypothesis.

The ${\rm Decoder}$ consists of a multi-layer LSTM while ${\rm DecoderOut}$ consists of an affine transform with a $Softmax(\cdot)$ output layer. The model is trained to minimize the cross entropy loss $\mathcal{L}(\cdot)$
between prediction ${\bf Y}$ and reference label ${\bf R}=\{r_1,...,r_N, \langle eos\rangle\}$
 \begin{equation}
    \mathcal{L}({\bf Y}, {\bf R}) = -\sum_n {y_n \log(r_n)}
\label{eq:cross_entropy}
 \end{equation}

\section{Proposed Approach -- FTL Platform}
\label{sec:proposal}

The developed FTL platform simulates the FL training process but without further investigation of either the communication or the privacy and encryption aspects of the task.
Although a seq2seq ASR model~\cite{Chiu+18} is herein used as a test-case, the findings and conclusions of the proposed approach are generalizable to other tasks as well, such as Computer Vision, text processing, and edge computing.
As mentioned in Section~\ref{sec:intro}, the focus of this paper is: first, on optimizing the training strategy in terms of task classification performance and speed of convergence and, second, on finding ways of leveraging untranscribed data, like that found in end-user devices, with more details found in~\cite{KDGG20}.

\subsection{System Description}

\begin{figure}[ht]
  \centering
  \includegraphics[width=.9\columnwidth]{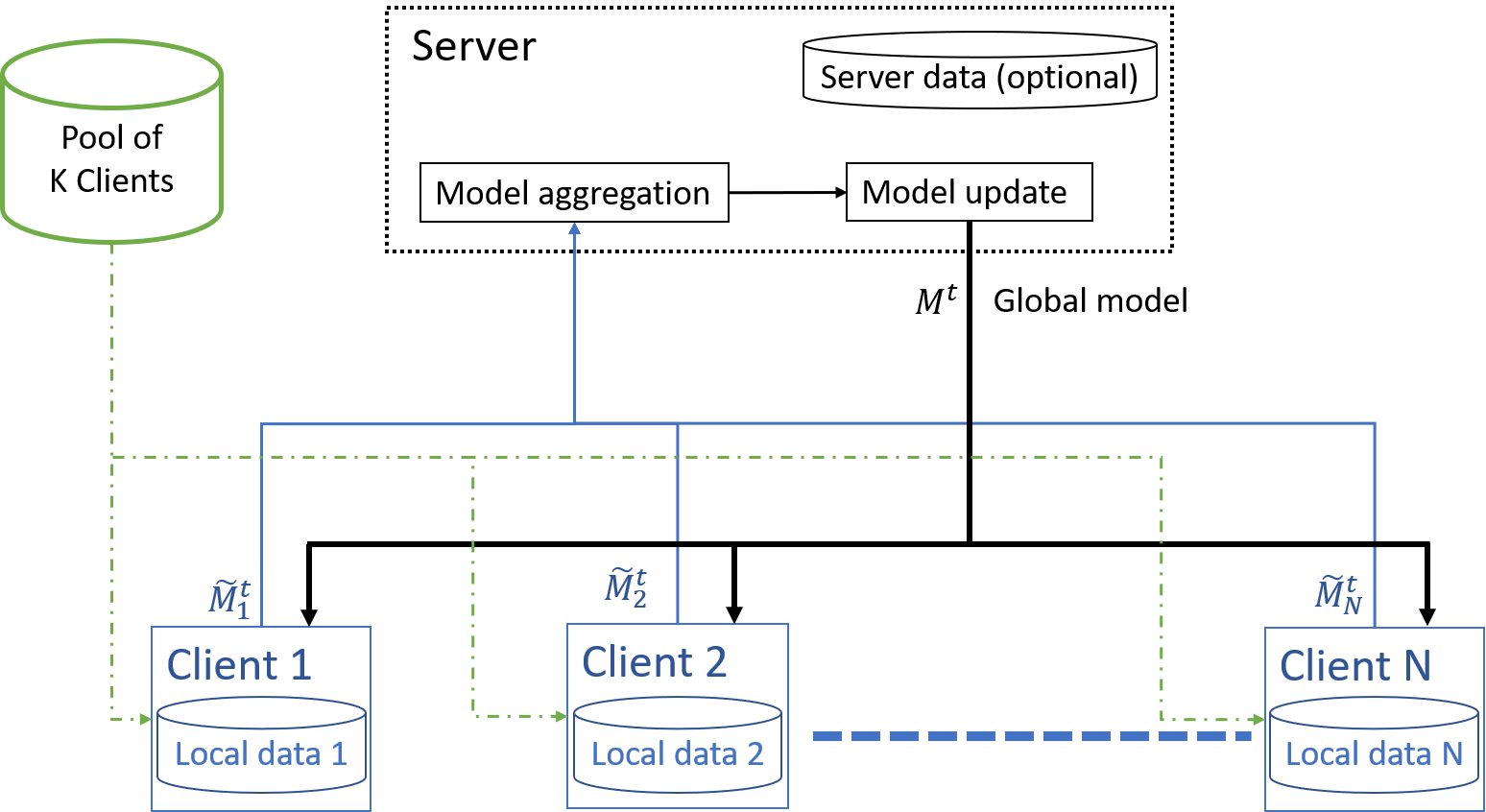}
  \caption{FTL Platform: Processing N clients per batch out of a pool of K clients.}
  \label{fig:FTL_platform}
\end{figure}

The proposed system, as depicted in Figure~\ref{fig:FTL_platform}, consists of a pool of $K$ (remote) clients with fixed datasets assigned to each client. Contrary to DT, the training data is not reshuffled after every epoch, but the initial data segregation setup is fixed throughout the task. Every iteration $t$ consists of processing randomly sampled $N \ll K$  clients, and returning them to the pool -- random sampling with replacement. The use of just these $N$ clients without loss in performance provides additional flexibility unique to the proposed FTL platform. Additionally, limiting the processing to the $N$ nodes decreases the latency between iterations and enhances the robustness against rogue nodes or attacks.

Once finished processing data for these $N$ clients, the updated models $\tilde{M}_T^{(j)}, \ j=1,..,N$ are aggregated, and a global gradient is estimated. This gradient is used to update the global model before the next iteration $T+1$, where $T$ depicts time instances on the server-side when the seed model is updated. Due to this sampling of the clients, the sweeping of all data takes longer. However, experimental results have shown that it is neither necessary for reaching an optimal point nor detrimental for the overall model performance.  

The FTL simulation platform allows for a realistic simulation of an FL system while excluding communication aspects such as encryption or rogue nodes. 
Due to the large size of our production-scale SR models, we have constrained the simulation platform to leverage multiple GPUs so that the training time remains in the order of hours or days instead of weeks or months\footnote{New models with $1/10^{th}$ of the original footprint are currently trained. Such models can be deployed on clients with memory constraints.}.
At the same time, we have designed the platform to support an arbitrary number of clients while utilizing a fixed and typically much smaller number of GPUs.
To address the requirements mentioned above, we have implemented the simulation as an MPI program with $P$ processes,~\cite{MPI_standard}, each of which has access to a dedicated GPU in a (potentially multi-node) GPU cluster.
The process with rank $0$ simulates the server while the remaining $P - 1$ processes are used as workers, responsible for simulating the clients' model training process.

In each iteration, the server randomly samples $N \ll K$ clients to participate in training, as described above.
The simulation platform executes model training for these $N$ clients by dispatching training ``assignments'' to the workers; an idle worker performs the training process of the specific client with all necessary parameters such as the global model, optimizer parameters and the identifier of the client-specific data set.  After performing training for that client, the worker sends the privacy-insensitive results, i.e., the locally trained model and training loss, back to the server process.
Then, the server updates the dictionary of the available resources and the clients yet to be processed.
This simple scheduling process continues until all $N$ clients have been processed.
To preserve memory on the server, the client models are aggregated in a streaming fashion as they are returned by the workers.
For example, in the case of a simple model averaging, the server aggregates all the client models into a single copy of the model that is stored in GPU memory and then divides the model parameters by $N$ to compute their average.
Once the aggregated model is computed, the server uses it for updating the global model and then proceed to the next iteration.

\subsection{Hierarchical Optimization}
\label{subsec:hier_optim}

Mini-batch optimization methods, extending classical stochastic methods to process multiple
data points in parallel, have emerged as a popular paradigm for FL~\cite{LSTS18}. Approaches like  ``Federated Averaging'' (FedAvg)~\cite{KMR15}, a method based on averaging local models after stochastic gradient descent (SGD) updates, is often considered as the golden-standard approach. FedAvg is shown to generalize well while significantly improving performance in terms of speed-ups. Lately, ``Generalized FedAvg'' was presented in~\cite{Reddi+20, Leroy+19}, a method with similarities to our proposed hierarchical optimization method.

Herein, a different approach is proposed, one of a hierarchical optimization process. The training process consists of two optimization steps: first, on the client-side using a ``local'' optimizer, and then on the server-side with a ``global'' optimizer utilizing the aggregated gradient estimates. The two-level optimization approach combines the merits of FedAvg with additional speed-ups due to the second optimizer on the server side. Further, aggregating the gradient estimates is shown to be beneficial since more data per iteration is included. The proposed algorithm is shown to converge faster than the centralized training method implemented on Horovod or even BMUF, after data volume normalization~\footnote{The comparison is in terms of iterations required for convergence}. An overview of the proposed algorithm is shown in Alg.~\ref{alg:Hier_Optim}.

In more detail, the $j^{th}$ client update runs $t$ iterations with $t\in[0, T_j]$, locally updating the seed model $w^{(s)}_T$ with (herein shown using the SDG optimizer, without loss of generality) with a learning rate of $\eta_j$,
\begin{equation}
    w^{(j)}_{t+1}=w^{(j)}_t - \eta_j \nabla  w^{(j)}_t
\label{eq:local_update}
\end{equation}
where $t$ is the local iterations on $j^{th}$-client, i.e., depicts the client time steps, and  $w_t^{(j)}$ the local model and $w_0^{(j)}\defeq w^{(s)}_T$.

The $j^{th}$ client returns a smooth approximation of the local gradient $\tilde{g}_T^{(j)}$ (over the $T_j$ local iterations and $T$ is the iteration ``time'' on the server side) as the difference between the latest, updated local model $w_{T_j}^{(j)}$ and the previous global model $w^{(s)}_T$ 
\begin{equation}
   \tilde{g}_T^{(j)} = w^{(j)}_{T_j} -   w_T^{(s)}
    \label{eq:local_gradient}
\end{equation}
Since, estimating the gradients $g_T$ is extremely difficult, hereafter the approximation $\tilde{g}_T^{(j)}$ is used instead.

The gradient samples $\tilde{g}_T^{(j)}$ are weighted and aggregated, as described in Section~\ref{sec:aggregate}
\begin{equation}
   g^{(s)}_T =  \sum_j{\alpha_T^{(j)} \tilde{g}_T^{(j)}}
\label{eq:global_grad}
\end{equation}
where $\alpha_T^{(j)}$ are the weights for the aggregation step, as described in Section~\ref{sec:aggregate}.

The global model $w^{(s)}_{T+1}$ is updated as in~(\ref{eq:global_grad}) (here also shown using SGD, although not necessary), \begin{equation}
   w^{(s)}_{T+1} = w_T^{(s)} - \eta_s g^{(s)}_T
    \label{eq:global_update}
\end{equation}
where $\alpha_T^{(j)}$ are the weights for the aggregation step, as described in Section~\ref{sec:aggregate}.

The process described in Equation~\ref{eq:server_training} is a form of ``\emph{Online Training}''~\cite{Parisi+18, SPLH17}.  While updating,  the seed model is drifting further away from the original task. In order to ensure compatibility with previous tasks, we propose a training step over held-out data (matching to the tasks in question) on the server side, Equation~\ref{eq:server_training}, after the model aggregation and update. This way, the model updates are regularized in a direction matching the held-out data. A ``gentle'' update of the model can avoid diverging too much from the task of interest. This is particularly useful for the case of imbalanced and/or vastly heterogeneous data. 
\begin{equation}
    w^{(s)}_{\tilde{t}+1} = w^{(s)}_{\tilde{t}} - \eta_w \nabla  w_{\tilde{t}}
    \label{eq:server_training}
\end{equation}
This training step on the server-side can be seen as an example of ``\emph{Naive Rehearsal},'' replaying previously seen data and ensuring ``backward'' compatibility.

\begin{algorithm}[htb]
\caption{Hierarchical Optimization}
\begin{algorithmic}[1]

\Procedure{HierOptim}{$w_0^{(s)},\ \bf{x}^{(j)}_T$}         
    \While{Model $w_T^{(s)}$ hasn't converged}         
        \For{$j\ in\ [0,N]$}
            \State Send seed model $w_T^{(s)}$ to $j^{(th)}$ clients 
            \State Train local models $w^{(j)}$ in $j^{th}$-node with data $x_T^{(j)}$
            \State Estimate a smooth approximation of the local gradients $\tilde{g}_t^{(j)}$ for $j^{th}$-node
            \State $j \leftarrow j+1$
        \EndFor
        \State Estimate weights $\alpha^{(j)}_T$ per DGA algorithm
        \State Aggregate weighted sum of gradients, Equation~\ref{eq:global_update}
        \State Update global model $w_T^{(s)}$ using aggregated gradient 
        \State Update global modal on held-out data, Equation~\ref{eq:server_training}
    \EndWhile
\EndProcedure
\end{algorithmic}
\label{alg:Hier_Optim}
\end{algorithm}

The convergence speed of training due to  the hierarchical optimization scheme is improved by a factor of $2\times$, without any negative impact in performance. Also, the communication overhead is significantly lower since the models are transferred twice per client and iteration (instead of transmitting the model gradients after every mini-batch, as in~\cite{McMahan+17}).

\subsection{Unsupervised Training}

Accurate labels are not always available in many FL scenarios and SR applications. In such cases, efficient unsupervised training is crucial. In this work, we employ two unsupervised training methods, either utilizing multiple hypotheses~\cite{KDGG20} or based on available local text. In more detail, the first algorithm processes the $N$-best hypothesis of the speech recognizer as a sequence of soft labels. It is shown that such an $N$-best hypothesis, even with $N$ relatively small, has coverage of around $90\%$ of the correct labels. As such, the network is updated with the soft-labels as part of multi-task training, where the loss of each task is weighted based on the DGA algorithm, described in Section~\ref{sec:aggregate}. By doing so, we can alleviate degradation caused by determining a wrong hypothesis as the ground truth in contrast to the conventional SR adaptation techniques~\cite{HWG16}. We can also avoid a sub-optimal sequence-level solution encountered in semi-supervised training on the hybrid HMM system~\cite{LiSWZG17,MosnerWRPKSMH19}. Moreover, each task can be reasonably weighted based on the reliability of the hypothesis. 

The second approach is to adapt the model with TTS-based audio hierarchically. A mix of audio from TTS and randomly sampled speech is used as input for this approach. First, the seq2seq model is adapted with organization-level relevant data, and then, it is adapted with session-specific data in a federated manner. The first step creates a new seed model with the decoder lightly matching the expected session-based data. This seed model is used as the starting point for running the FTL pipeline described above on TTS-data per session. The TTS-based data causes the seq2seq model to diverge from the original one significantly, so the randomly sampled `real' data is used to regularize the training process, as in Equation~\ref{eq:server_training} - a phenomenon described above. A combination of both of the aforementioned methods as part of the processing pipeline is now investigated.

\section{Dynamic Gradient Aggregation}
\label{sec:aggregate}

Training with heterogeneous data poses additional challenges, especially for the aggregation step, as in Equation~\ref{eq:global_update}. Amongst these challenges are: 
\begin{enumerate*}[label=\roman*.]
\item \emph{data heterogeneity}: not all client data distributions are adequately captured by the model; thus, the corresponding training losses are expected significantly higher. In such cases, the model tends to move to a direction that is largely different from the rest of the gradients. 
\item \emph{data quality}: the quality of a particular local data partition might be quite different from the rest, leading to noisier gradients. 
Higher values for $\mathcal{L}^{(j)}$ coefficients can be seen as an indication of batches that are not well represented by the model. Possible sources of such loss values are either data of bad quality, e.g., noisy data, or data distributions further apart from the model. 
\item \emph{Adversarial or Byzantine attacks}: Especially Byzantine nodes~\cite{BMGS17} can create a similar situation with gradients very different than the expected.  
\end{enumerate*}
Either way, the model will be forced to drift further apart from the rest of the models. As a result,  the aggregation process will become noisier due to this diversification of the aggregated models. Batches,\footnote{We interchangeably use `batch' and `client data' in this analysis -- each client can be seen as a large batch of data, while FL training. The same algorithm can be used in DT where the `batch' coincides with a sequence of mini-batches.} where the training losses are of similar magnitude, would be expected to move the model in a similar direction; thus the aggregation process will be better aligned. Such alignment of the aggregated gradients is also beneficial for the convergence speed, as shown in the experimental section below. 

In any of the aforementioned scenarios,  gradients that deviate from the rest should be processed differently. 
Herein, the proposed approach is using weights during the aggregation step, i.e., by weighting the local gradients, $\tilde{g}_T^{(j)}$, in~(\ref{eq:local_gradient}), the contribution of some components can be de-emphasized. The proposed algorithm is called ``\emph{Dynamic Gradient Aggregation}'' (DGA). Two different flavors of DGA  are herein proposed: first, the  ``deterministic'' one using the training losses as the weighting coefficients, like in~\cite{Alain+16, BTPG16}, and the ``data-driven'' approach, where a neural network is trained to infer the weights.
In some tasks,  weighted aggregation does not significantly affect the overall WER performance (at least on the LibriSpeech task, where data is more homogeneous), however it makes the training convergence significantly faster. 
On the other hand, the DGA algorithm approach can significantly affect the convergence speed and classification performance in either unsupervised training where the label quality can vary significantly or in  very diverse local data found in FL scenarios.  
The weighting process can be seen as a type of regularization, de-emphasizing gradient directions, where the local models can diverge too much. Thus, the back-propagation updates are based on less noisy mini-batch gradients.  

The ``deterministic'' approach or ``Softmax Weighting'' (or SM\_DGA) utilizes the negative training loss coefficients $\mathcal{L}$, from Equation~\ref{eq:cross_entropy}, as weights $\alpha^{(j)}_t \sim \mathcal{L}(\cdot)$. These weights are normalized when passed through a $Softmax(\cdot)$ layer,
\begin{equation}
\alpha^{(j)}_T=\exp(\mathcal{-\beta L}^{(j)}_T)/\sum_i{\exp{(-\beta \mathcal{L}^{(i)}_T})}
\label{eq:softmax_weights}
\end{equation}
where $\beta$ is the temperature of the Softmax function. The temperature can regulate how aggressive can be the weighting of the gradient components. According to Equation~\ref{eq:softmax_weights}, the aggregation weights are smaller for those nodes, i.e. gradients, with larger values of the corresponding losses.

The second approach is based on Reinforcement Learning. The weights are inferred by a network, trained with rewards according to sparse, time-delayed labels.  This approach is called ``RL Weighting'' (or RL\_DGA).
An agent perceives a stimulus from the environment, called ``observation.'' The rewards used for training the agent depend on how good the agent's action is. Usually it is based on a specific, predefined reward policy. This agent takes action in order to optimize the interaction with the environment according to such rewards policy, while inducing new states to the system. Then, updated observations and a new reward are acquired based in  such new state. Herein, our approach is based on Active  Reinforcement Learning, where the reward depends directly on the action selected~\cite{EVD08}.  

By leveraging Reinforcement Learning, a neural network $RL(\cdot)$ (or equally, agent) is used for inferring the weights based on a set of input features $\bf{x}_T^{(j)}$ (or equally observations) from each of the $j$-clients. The agent decides on the values of the gradient weights aiming at improving the CER performance of the model. We propose training an end-to-end system that takes as input the training loss coefficients and gradient statistics and learns the optimal weighting strategy.

\begin{algorithm}[htb]
\caption{Dynamic Gradient Aggregation Based on Reinforcement Learning}
\begin{algorithmic}[1]

\Procedure{RL\_DGA}{$\tilde{g}^{(j)}_T, \bf{x}^{(j)}_T$}         
    \State $RL(\cdot)$ Model Initialization
    \While{Model $w_T^{(s)}$ hasn't converged}         
        \State Read the Observations $\bf{x}^{(j)}_T$
        \State Estimate Weights $a^{(j)}_T \leftarrow  RL(\bf{x}^{(j)}_T) $
        \State Estimate Weights $b^{(j)}_T \leftarrow Softmax(\bf{x}^{(j)}_T)$, from Equation~\ref{eq:softmax_weights}
    
        \State Estimate $\tilde{W}^{(s),a}_T$ and $\tilde{W}^{(s),b}_T$ based on two sets of aggregation weights $a^{(j)}_T, \  b^{(j)}_T$ with Equation~\ref{eq:global_update}
    
        \State Estimate $CER_{a, T},\ CER_{b, T}$ for the new models $\tilde{W}^{(s),a}_T, \tilde{W}^{(s),b}_T$     
        
        \If {$CER_{a, T}- CER_{b, T}>\theta$}
            \State $w^{(s)}_{T+1} \leftarrow \tilde{w}^{(s),a}_T$
            \State $r_T=R$
        \ElsIf{ $ abs(CER_{a, T}- CER_{b, T}) \leq \theta$ }
            \State $w^{(s)}_{T+1} \leftarrow \tilde{w}^{(s),a}_T$
            \State $r_T=0.1R$
        \Else
            \State $w^{(s)}_{T+1} \leftarrow \tilde{w}^{(s),b}_T$
            \State $r_T=-R$
        \EndIf
        \State Update model $RL(\cdot)$ on reward $r_T$ 
    \EndWhile
\EndProcedure
\end{algorithmic}
\label{alg:RL_Training}
\end{algorithm}

\begin{figure}[t]
  \centering
  \includegraphics[width=.7\columnwidth]{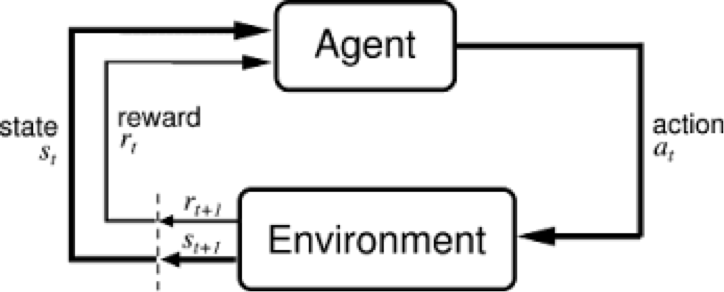}
  \caption{Simple Reinforcement Learning schema}
  \label{fig:RL_schema}
\end{figure}

Translating the training of the network is terms of  Reinforcement Learning (RL), the ``actions'' are predicted from the results they incur, i.e., the ``Character Error Rate'' (CER) on the validation set in every time step $T$. The RL network learns the sequence of actions that lead the ``agent'' to maximize its objective function (or policy). This reward policy is based on the CER performance, i.e., the ``environment,'' the ``action,'' and ``state'' $s_{T+1}$ is the new aggregated model, as in Figure~\ref{fig:RL_schema} from~\cite{SuBa18}. 
In more detail, the ``environment'' in every iteration $T$ is the gradient components $\tilde{g}_T^{(j)}$, the states $s_T$ are described by the input features $\bf{x}^{(j)}_T$, and the action vector $\alpha^{(j)}_T$ is the aggregation weights. The policy is dictated by the output CER performance (on the validation set) of the updated model~(Equation~\ref{eq:global_update}). The policy agent is a DNN trained as detailed. An overview of the algorithm can be found in Alg.~\ref{alg:RL_Training}.

The reward policy is based on the CER performance of two different networks that are trained with the aggregated gradients~(\ref{eq:local_gradient}). These two networks are versions of the  same seed model, after training with either the inferred weights or the $Softmax$-based ones, Equation~\ref{eq:global_update}. Depending on the comparative results, a reward $r_T$ is provided, and the new state $s_{T+1}$ is estimated. The threshold and reward parameters $\theta,\ R$ are part of the reward policy as detailed in Alg.~\ref{alg:RL_Training}.
Herein, the input features $\bf{x}_T^{(j)}$ are the combination of the training loss coefficients (as described above) augmented by the gradient magnitude mean and variance values. The gradient-related coefficients are estimated over all the parameter gradients during the local (i.e., on the $j^{th}$-client) training iterations.  

Unlike other machine learning paradigms, Reinforcement Learning does not require supervision, but just a reward signal. 
Also, the feedback can be delayed: It does not have to be instantaneous as in supervised learning algorithms. 
Data is sequential, and the agent actions affect only the subsequent data it receives (i.e., exploitation approach). 
Since this approximation is unstable, a replay memory is introduced. Random mini-batches from the replay memory are used instead of using the most recent transition. 
This breaks the similarity of subsequent training samples, which would otherwise drive the neural network into a local minimum, prematurely ending the training process.

\section{Experiments and Results}
\label{sec:experiments}

Two datasets are used as the experimental test-beds, the LibriSpeech task~\cite{PCPK15} (LS task) for supervised training, and an internal dataset based on Powerpoint presentations for the unsupervised task. The first dataset contains about $1k$ hours of speech from $2.5k$ speakers reading books. The second setup is based on $2.8M$ written sentences (in detail, $32M$ words with a dictionary size of $172k$ unique words) of internal documents and emails from Microsoft employees. This written corpus is pushed through the Neural TTS Microsoft service, creating $3.5k$ hours of audio. In this experiment, a 6-layer bLSTM, with dropouts, is used for the encoder, 2 layers of uni-directional  LSTM are used for the decoder, and finally, a conventional location-aware content-based attention layer with a single head is used. The input features are 80-dim log mel filter-bank energies, extracted every $10\ msec$, 3 consecutive stacked frames are set as input, and $16k$  subwords based on a  unigram language model are used as the recognition unit. For the first scenario, the baseline model is a state-of-the-art seq2seq model trained using Horovod on the entire training set of $920h$. This model's performance provides the lower bound of the WER for the particular Speech Recognition task since all the data is used in a centralized manner. For the $2^{nd}$ scenario, a LAS model of similar architecture is used as the seed model. This later model is trained on $75k$ hours of speech.

\subsection{Supervised Training FTL Experiments}
\label{subsec:ftl_experiments}

For the supervised experiments, the training is split into two parts of 460h each, with no overlapping speakers. The first part is used to train a seed model, without ever using this data again. Then, the 2nd part of the dataset is used to simulate online training, under the FL conditions. We will follow two different directions for the FL training process: first, the training set is split into 7 distinct parts, never reshuffling the data again (contrary to DT approaches). These data splits are random, with no overlapping speakers across them. The second direction is to segregate the data based on the $1100$ speaker labels. Each one, either $K=7$ or  $K=1.1k$, of the partitions, is assigned to a client. In the FTL framework, all clients are unaware of the rest of them -- only the server ``knows'' which of the clients are used, randomly sampling which ones will be aggregated. The number of sampled clients $N$ in our experiments varied from 25 to 400, with higher $N$ being better but with small fluctuation in overall performance. Based on the compromise between communication overheads and memory usage, we henceforth set $N=100$.

The weighting approach for $\alpha_T^{{j}}$ is either based on the training loss, herein noted as ``\emph{Softmax-weighting},'' or inferred by the $RL(\cdot)$ network,  stated as ``\emph{RL Weighting}.'' The $RL$ network is a 5-layer DNN with ReLU activations, and a bottleneck layer ($2^{nd}$ layer to last). The input layer size is $3N$, and the output layer $N$. The reward policy is $+1$ if the weights provide better CER value compared to the loss-based weights, $-1$ in the opposite case, and $+0.1$ when the performance of the two cases are similar. The network has a memory of the previous $1000$ instances, and it samples a mini-batch of $32$ instances for training per iteration.
\begin{table}[t]
	\centering
	\begin{tabular}{|c|l|c|}
	    \hline
		\multicolumn{3}{|c|}{LibriSpeech Task}\\ \hline\hline
		                            & Training Scenario & WER (\%)\\ \hline
		\multirow{3}{*}{Centralized }   & SotA (lower bound) & 4.00\% \\ \cline{2-3}\cline{2-3}
		                            & Training on 1st 50\% of LS(seed) & 5.66\% \\ \cline{2-3}
		                            & Online training on 2nd 50\% of LS & 4.61\% \\ \hline
		\multirow{5}{*}{FTL}        & FedAvg                        & 4.55\% \\ \cline{2-3}
		                            & Hier.  Optim. (7 clients)     & 4.51\% \\ \cline{2-3}
		                            & Hier.  Optim. (1.1k clients)  & 4.45\% \\ \cline{2-3}
		                            &   +  Softmax DGA              & 4.41\% \\ \cline{2-3}\cline{2-3}
		                            &   +   RL DGA                  & 4.40\% \\ \hline
	\end{tabular}
	\caption{System evaluation on the LibriSpeech task, for training offline seq2seq with attention model.}
    \label{tab:LS}
\end{table}

The top 3 rows in Table~\ref{tab:LS} are with centralized training, with the lower bound in performance coming from the model trained on the entire dataset (offline training). The $4\%$ WER for this model appears inline with the literature. The $2^{nd}$ model (``online training'' row) is based on the seed model initially trained on the 1st half of the data till convergence. The model is online trained with the $2^{nd}$ half of the data. In both steps, Horovod is utilized. 

Then, the second scenario simulating the FL condition is examined, where the seed model is further refined in an FL fashion by training on unseen data. 
Different strategies for model aggregation were investigated, such as model averaging (``FedAvg'' row in the Table), or hierarchical optimization using optimizers such as Adam, LAMB, LAR, and SGD. For the FedAvg system, the model averaging is performed on the server, while the SGD optimizer is used for training on the client's side. Combinations of the server/client optimizers were also investigated. The differences in performance of these combinations of optimizers were rather limited, and for the sake of space, are not further elaborated here. However, a state-less optimizer on the client-side is adopted as a standard, because the initial model is changing after each iteration\footnote{As discussed in Section~\ref{sec:proposal}, the server aggregates the client models and updates the seed (server) model. Then, this model is re-iterated to the clients.} and therefore, keeping the state of the previous iteration/model as part of the local optimizer didn't make much sense. Herein, the combination used for all the experiments is Adam/SGD for the server/client sides. SGD with momentum was also investigated without much difference in performance, though.

The next experiment was to transition to the per-speaker partitioning of the data, i.e., $1100$ partitions and an equal number of clients. As mentioned above, $100$ clients per iteration are sampled out of the pool of $1100$ and finally aggregated. The transition from a homogeneous data split, i.e., the case of 7 partitions, to a more heterogeneous per-speaker partition, improved the overall model performance lightly. This can be explained by the additional diversity provided when aggregating such client models. However, due to this diversity, the convergence during training required additional iterations. 

The proposed DGA algorithm has addressed this issue by de-emphasizing the gradients from clients loosely modeled. The FedAvg system requires around 800 iterations for convergence. Such a system is too slow, and henceforth, it will not be considered as the state-of-the-art baseline. Our baseline system is based on hierarchical optimization but without DGA. Even though the overall performance was not impacted (this is only true for the case of the LS task), the overall convergence speed was improved by a factor of  $~1.5\times$ compared to the baseline system and $~7\times$ compared to the FedAvg system. In more detail, approximately 384 iterations are required for the case of unweighted aggregation against only 224 for the SM\_DGA. In the case of RL\_DGA, the number of required iterations is even lower, decreased by an additional factor of $1.5\times $, requiring only 144 iterations. The variations in performance between different approaches are limited; however, the task is quite homogeneous. Further improvements in other in-house tasks, e.g., adaptation on presentation sessions, have also been realized.  

\begin{figure}[t]
  \centering
  \includegraphics[width=1.025\columnwidth]{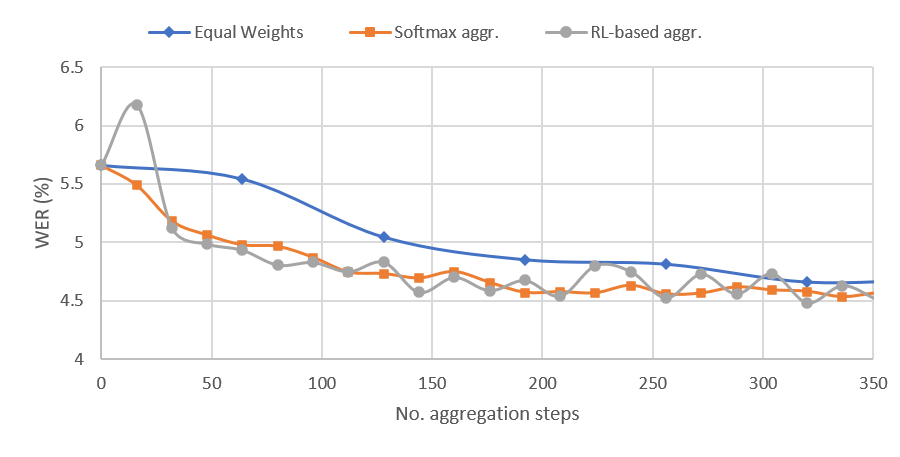}
  \caption{Training on LibriSpeech corpus for 3 weighting scenarios, i.e., uniform weights, Softmax- and RL-based weights.}
  \label{fig:iterations}
\end{figure}
The convergence plots for the different weighting scenarios, i.e., uniform weights, Softmax- and RL-based weighting, for the LS task is shown in Figure~\ref{fig:iterations}. 
The RL-based aggregation curve shows ripples (particularly wide at the beginning of the training process), but this is expected since the RL network starts from a random state and convergences later on. Further, the performance of the RL- and Softmax-based weighting schemes seem to converge after a few iterations. This is also expected since the rewards policy is based on the Softmax- performance. The RL network learns Softmax-based behavior after a while. However, it is consistently outperforming that system in terms of convergence speed. The LS dataset is quite homogeneous in terms of audio quality, so the proposed algorithm is not expected to perform vastly different in terms of WER performance. However, the RL-based system outperforms the uniform-weighted one in other tasks with improvements up to $11\%$ WERR.

\subsection{Hierarchical Unsupervised Training with Federated Session Adaptation}
\label{subsec:unsupervised_experiments}

In the case of unsupervised training, our approach is to adapt the seq2seq models hierarchically: first, adapt the seed model to organizational-level tenant data, and then perform session adaptation based on the FTL platform. The dataset used here consists of TTS data and a random sample of real speech already used for training the seq2seq model; the use of the real data prevents the model from overfitting to the TTS data. The test set is based on four presentations of about $40min$ each, with 1230 sentences, $12k$ words, and a vocabulary size of $2.5k$ words. The text data found in the presentation slides were also used to synthesize the speech data (using the TTS engine as described below) with a total size of $1.3h$. As an initial model, the LAS model trained on $75k$ hours of data is used. The seed model is adapted with the TTS data generated from the tenant text data in a centralized way. Several scenarios for this adaptation step are investigated, performing subspace adaptation, i.e., changing the encoder, decoder or both, and including different sources of the TTS speech data: the tenant TTS data only, tenant and presentation slide TTS data, and the mixture of the TTS and real-speech data. The final adapted model matches the content of the presentation slides better without the exact transcript of the presentation speech.

Once the seed model is adapted to the tenant text data, it is now used as the starting model for the second adaptations step. This second adaptation step is based on the FTL platform, where the 4 meetings and the new model are used for the final inference step. 

The new model is adapted iteratively to the presentation-related text using the TTS-based audio. The DGA step is also applied to weight the input gradients accordingly. However, we noticed that the model overfits very fast on the synthetic TTS data, with the overall performance steeply deteriorating. To address this issue, we have added real-speech data on the server-side training to regularize the process. The real-speech audio is randomly picked from the training set. Note here that this random subset of speech has already been used for training the initial LAS model -- no need for held-out data. The addition of this set reduces the model drift significantly while improving the overall recognition performance. As mentioned above, this step resembles the ``Naive Rehearsal'' approach alleviating any Catastrophic Forgetting effect.
\begin{table}[t]
	\centering
	\begin{tabular}{|c|l|c|c|}
	    \hline
		\multicolumn{4}{|c|}{Presentation-based Session Adaptation}\\ \hline\hline
		                                &Training Scenario  & Adapt. Comp.  & WER (\%)\\ \hline
		\multirow{5}{*}{Centralized }   & Baseline          & None          & 6.86\% \\ \cline{2-4}\cline{2-4}
		                                & \multirow{2}{*}{Tenant Text}  & Encoder   & 8.71\% \\ \cline{3-4}
		                                                                && Decoder  & 6.41\% \\ \cline{2-4}
		                            & \  + PPT-audio + DGA              & Decoder  & 6.46\% \\ \cline{2-4}
		                            & \ \ \ \ \ \ + real-speech            & Decoder  & 6.29\% \\ \hline\hline
		FTL                         & TTS- and real-speech              & Decoder  & 5.51\% \\ \hline\hline
	\end{tabular}
	\caption{Hierarchical Unsupervised Training with Federated Session Adaptation.}
    \label{tab:PPT}
\end{table}

The hierarchical approach shows an $~20\%$ WERR improvement over the original model performance. We have investigated TTS adaptation on the speakers' voice (each of the 4 presentations is assumed to contain a single speaker), but no additional benefits in performance were found.

\section{Discussion and Future Work} 
\label{sec:conclusion}

In this work, a novel Federated Learning platform for Speech Recognition tasks is presented. This is the first of its kind as far as the authors know. Herein, Federated Learning approaches for other tasks were investigated and compared with the proposed ones. Although the discussion about the platform is focused on the ASR tasks in hand, the FTL platform can be easily generalized to other tasks, such as FaceID or NLU-related. Currently, we are working on other classification tasks using the FTL platform, employing other modalities.

In addition to those approaches, we are presenting novel algorithms addressing challenges unique to the Speech Recognition scenario. This novel approach of weighting the gradients between mini-batches allows for enhanced convergence speed-ups and improved model performance.  The proposed gradient aggregation scheme acts as a regularizer de-emphasizing batches where the data are not well modeled.  
Herein, a weighted gradient aggregation algorithm is described enabling $7\times$ speed-up and $6\%$ WERR on LibriSpeech task and $20\%$ WERR for a session adaptation task. 

\section{Acknowledgements}
The authors would like to thank Masaki Itagaki, Ziad Al Bawab, Lei He, Michael Zeng, Xuedong Huang, Veljko Miljanic and Frank Seide for their project support and technical discussions.


\ninept
\bibliographystyle{IEEEbib}
\bibliography{refs}

\end{document}